\documentclass[10pt,twocolumn,letterpaper]{article}

\usepackage[pagenumbers]{cvpr} % To force page numbers, e.g. for an arXiv version

\usepackage{cite}
\usepackage{amsmath,amssymb,amsfonts}
\usepackage{algorithmic}
\usepackage{graphicx}
\usepackage{textcomp}

\usepackage{pifont}
\usepackage{booktabs}
\usepackage{algorithm}
\usepackage{textcomp}
\usepackage{multirow}
\usepackage{hhline, booktabs}
\usepackage{xcolor}
\usepackage{soul}   % For highlighting
\usepackage{colortbl}
\usepackage{float} % for forcing the placement of figures
\usepackage{caption} % for custom captions
\usepackage{url}
\usepackage{enumerate}
\usepackage{tikz}
\usepackage{adjustbox}
\usepackage{comment}
\usepackage{scalefnt}

\newcommand{\xmark}{\ding{55}}%
\newcommand{\cmark}{\ding{51}}%

\definecolor{MyColorTab}{RGB}{230, 230, 230}
\definecolor{ForestGreen}{RGB}{34,139,34}
\definecolor{OrangeRed}{RGB}{236,83,83}

\usepackage{lipsum}

\newlength{\RoundedBoxWidth}
\newsavebox{\GrayRoundedBox}
   {\setlength{\RoundedBoxWidth}{\dimexpr#1}
    \begin{lrbox}{\GrayRoundedBox}
       \begin{minipage}{\RoundedBoxWidth}}%
   {   \end{minipage}
    \end{lrbox}
    \begin{center}
    \begin{tikzpicture}%
       \draw node[draw=black,fill=black!10,%
             inner sep=2ex,text width=\RoundedBoxWidth]%
             {\usebox{\GrayRoundedBox}};
    \end{tikzpicture}
    \end{center}}

\usepackage[most]{tcolorbox} % for colored boxes

\tcbset{
  llmprompt/.style={
    colback=blue!3!white,     % background color
    colframe=blue!40!black,   % border color
    left=4pt,
    right=4pt,
    top=4pt,
    bottom=4pt,
    boxrule=0.5pt,
    arc=2pt,                  % slightly rounded corners
    outer arc=2pt,
    width=\columnwidth,       % match IEEE column width
  }
}

\usepackage{tikz}
\usepackage{lipsum}

\definecolor{cvprblue}{rgb}{0.21,0.49,0.74}
\usepackage[pagebackref,breaklinks,colorlinks,citecolor=cvprblue]{hyperref}

\def\paperID{*****} % *** Enter the Paper ID here
\def\confName{CVPR}
\def\confYear{2024}

\title{ViConEx-Med: Visual Concept Explainability via Multi-Concept Token Transformer for Medical Image Analysis}

\author{Cristiano Patrício\textsuperscript{1,2,4}\thanks{Correspondence to: \texttt{cristiano.patricio@ubi.pt}}, Luís F. Teixeira\textsuperscript{3,4}, João C. Neves\textsuperscript{1,2}
\\
\small\textsuperscript{1}Universidade da Beira Interior
\quad\textsuperscript{2}NOVA LINCS
\quad\textsuperscript{3}Faculdade de Engenharia da Universidade do Porto
\quad\textsuperscript{4}INESC TEC
}

\begin{document}
\maketitle

\begin{abstract}
Concept-based models aim to explain model decisions with human-understandable concepts. However, most existing approaches treat concepts as numerical attributes, without providing complementary visual explanations that could localize the predicted concepts. This limits their utility in real-world applications and particularly in high-stakes scenarios, such as medical use-cases. This paper proposes ViConEx-Med, a novel transformer-based framework for visual concept explainability, which introduces multi-concept learnable tokens to jointly predict and localize visual concepts. By leveraging specialized attention layers for processing visual and text-based concept tokens, our method produces concept-level localization maps while maintaining high predictive accuracy. Experiments on both synthetic and real-world medical datasets demonstrate that ViConEx-Med outperforms prior concept-based models and achieves competitive performance with black-box models in terms of both concept detection and localization precision. Our results suggest a promising direction for building inherently interpretable models grounded in visual concepts. Code is publicly available at \url{https://github.com/CristianoPatricio/viconex-med}. 
\end{abstract}

\section{Introduction}
\label{sec:introduction}

\begin{figure}[!t]
    \centering
    \includegraphics[width=0.9\linewidth]{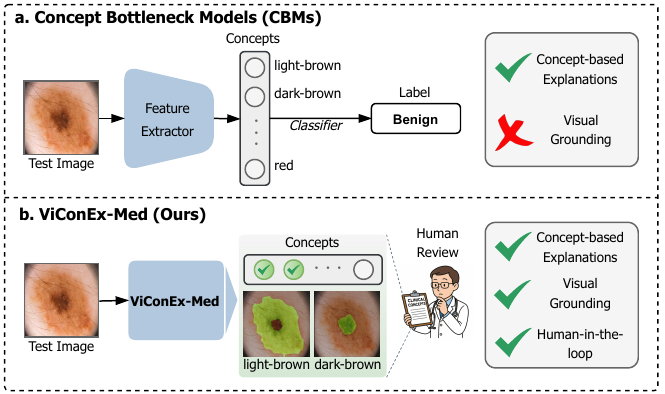}
    \caption{\textbf{(a)} Concept Bottleneck Models (CBMs) provide concept predictions but lack spatially-grounded visual explanations. \textbf{(b)} \textbf{\textit{ViConEx-Med}} produces faithful visual explanations for predicted concepts, enabling human-in-the-loop decision-making and fostering trust in clinical practice.}
    \label{fig:viconex-med_teaser}
    \vspace{-0.5em}
\end{figure}

Deep learning models have revolutionized medical image analysis by their ability to automatically learn discriminative features from complex imaging data, enabling efficient and accurate diagnosis across various disease detection tasks~\cite{li2023medical}. The use of sophisticated architectures, such as Convolutional Neural Networks (CNNs) and Vision Transformers (ViTs) have substantially improved performance in a wide range of medical imaging applications, including skin lesion classification where some approaches achieve diagnostic precision comparable to that of expert dermatologists~\cite{esteva2017dermatologist, codella2017deep}. Despite these advances, current state-of-the-art deep learning approaches remain largely opaque, often functioning as ``black boxes'' that provide predictions without human-interpretable reasoning. This lack of transparency poses a significant barrier to clinical integration, as explainability is essential for establishing reliability and supporting informed decision-making in high-stakes medical contexts~\cite{rotemberg2019role}. To address this issue, various approaches have been proposed to demystify the inner workings of deep neural networks (DNNs) ~\cite{patricio2023explainable}. Among these, approaches based on Concept Bottleneck Models (CBMs)~\cite{koh_2020_ICML} have gained substantial attention in the field of medical image analysis and eXplainable Artificial Intelligence (XAI)~\cite{Fang_MM2020, Lucieri_IJCNN2020, Patricio_2023_CVPRW, CLAT, bie2024mica}. These models introduce an intermediate bottleneck layer between the backbone and the classification head, ensuring that decisions are based on the human-interpretable concepts predicted by the bottleneck layer. This design mimics the way clinicians assess medical images, thereby enhancing transparency and fostering trust in model outputs. 
While effective, CBMs require task-specific annotations of human-understandable concepts and target classes, a combination that is often scarce in most of medical image datasets. Moreover, they lack the capacity to provide spatially grounded visual explanations for the predicted concepts.
These limitations highlight the need for inherently interpretable concept-based models that not only predict the presence of concepts but also provide visual explanations that localize these concepts within the image with minimal supervision. Recent methods~\cite{Patricio_2023_CVPRW, bie2024mica, CLAT} have attempted to combine concept prediction with visual explanation to improve transparency and interpretability of model decisions. However, most of these methods struggle to generate faithful and concept-discriminative visual explanations that directly correspond to the predicted concepts, primarily due to the lack of specialized architectural mechanisms explicitly optimized for this purpose.
Moreover, training typically requires both image-level concept annotations and corresponding target labels, a combination that is often scarce in medical imaging datasets. This limitation underscores the need for methods capable of operating under weak supervision. In the context of skin lesion diagnosis, clinicians usually rely on the ABCDE rule~\cite{rigel2005abcde}, which evaluates features such as asymmetry, border irregularity, color variation, diameter, and evolution. Each feature is assigned a score, and the cumulative score determines whether a lesion is considered suspicious for melanoma. Inspired by clinical practice, we argue that an interpretable AI tool capable of predicting such features and providing plausible visual explanations can offer valuable decision support~\cite{yan2024general}. This is particularly relevant in primary care settings, where the tool could help triage images of suspicious lesions for urgent referral or reassurance~\cite{jones2025using}. This paradigm offers several benefits, including facilitating early diagnosis and supporting a human-in-the-loop approach, in which clinicians or healthcare professionals review every patient in tandem with the predictions provided by the interpretable AI model.

In this work, we introduce \textit{\textbf{ViConEx-Med}}, a multi-concept token transformer for interpretable concept prediction in medical imaging. Our model enables accurate visual explanations for each predicted concept by leveraging a ViT-based architecture with multiple learnable tokens, each targeted to represent a specific concept. To further enhance the intra-concept diversity, we introduce an additional text-guided concept branch that incorporates textual embeddings from a medical foundation model, enhancing global visual concept tokens with complementary language semantics. The two modalities, visual and textual concept tokens, are jointly optimized to correlate pixel features, thereby improving concept localization. Finally, to ensure that each concept token attends to distinct image regions, we employ a contrastive objective that enforces their distinctiveness at each dedicated transformer block. Importantly, \textit{\textbf{ViConEx-Med}} achieves these properties using only concept-level labels, which indicate the presence or absence of concepts (e.g., colors) without requiring pixel-level annotations or target class labels. During inference, the model takes a medical image as input, predicts the corresponding concepts, and generates concept localization maps that visually ground each prediction, as illustrated in Figure \ref{fig:viconex-med_teaser}.
While concept-level labels enable effective training, their utility in practice is often limited by the difficulty non-experts face in interpreting them. To address this, we also introduce \textbf{\textit{SynSkin}}, a synthetic skin lesion-like dataset with image-level color labels and pixel-level masks. This dataset serves both as a benchmarking dataset to validate our approach and as an augmentation data source to improve the performance on existing medical datasets.
To evaluate the performance of our method, six medical image datasets were selected based on the availability of annotations at the concept level, including dermatoscopic datasets and a fundus image dataset. Our contributions can be summarized as follows:
\begin{itemize}
\item We propose \textit{\textbf{ViConEx-Med}}, a novel multi-concept token transformer framework that learns concept-specific localization maps directly from concept-to-patch attentions of multiple concept tokens;
\item We introduce \textbf{\textit{SynSkin}}, a large-scale synthetic image dataset with multilabel color annotations and pixel-level masks, specifically designed to facilitate robust evaluation of explainable models;
\item We conduct extensive experiments on six medical imaging datasets, demonstrating that our framework jointly provides accurate concept predictions and faithful visual concept explanations, outperforming state-of-the-art baselines while achieving competitive performance with black-box models.
\end{itemize}

\section{Related Work}
\label{sec:related_work}

\paragraph{Multi-Class Token Vision Transformers.}
% General overview and existing works
Vision Transformers (ViTs) have exhibited remarkable performance in medical image classification~\cite{yu2021mil,manzari2023medvit} compared to traditional convolution-based models~\cite{khan2022transformers}. Standard ViTs typically rely on a single class token, known as \texttt{CLS}, to summarize the information from the entire sequence of patch tokens, which is then fed into a linear classifier. 
MCTformer~\cite{xu2022multi} introduced multiple class tokens to ViTs, enabling class-discriminative localization via class-to-patch attention. Subsequent works have extended this idea for medical imaging. Gao et al.~\cite{Gao_Aligning_MICCAI2024} used cross-attention to learn multiple concept tokens for skin lesion diagnosis. Wen et al.~\cite{CLAT} encoded each retinal lesion as a learnable token. MCTformer+~\cite{xu2024mctformerplus} extended MCTformer~\cite{xu2022multi} by adding a regularization term to encourage class token separation in the embedding space and by replacing global average pooling with weighted pooling, thereby emphasizing informative patches in the final prediction.
% The limitations of current approaches
Despite their strong performance, these methods face limitations. \cite{Gao_Aligning_MICCAI2024} struggles with concept-discriminative localization due to coarse annotations. CLAT~\cite{CLAT} lacks explicit token discrimination, and MCTformer variants are constrained by the CLS-token paradigm. CaiT~\cite{touvron2021going} addressed these issues by decoupling patch and class-attention layers and introducing per-channel residual scaling, improving the representation of patch embeddings.
% Our proposal
Building on these insights, our method employs a CaiT-like architecture that separates patch self-attention from visual and textual-concept token processing.  Additionally, we leverage a contrastive objective on the learnable concept tokens to enforce their distinctiveness, leading to improved concept discrimination and more precise localization (\emph{cf.} Figure \ref{fig:qualitative_results_color_datasets}).

\paragraph{Concept-Based Models.}
% General overview and existing approaches
Concept Bottleneck Models (CBMs)~\cite{koh_2020_ICML} offer inherent interpretability by grounding final predictions in intermediate human-understandable concepts. However, they require extensive manual annotations and often lack visual concept explainability. To reduce annotation burden, some works use Large Language Models (LLMs) to generate candidate concepts~\cite{yang2023language,oikarinen2023label,menon2022visual,yan2023robust}, while others leverage Vision-Language Models (VLMs) to align visual features with predefined concepts~\cite{patricio2024towards, kim2024transparent, Gao_Aligning_MICCAI2024}.
Several methods enhance CBMs with visual explanations. For example, Patrício et al.~\cite{Patricio_2023_CVPRW} guided concept filters with segmentation masks. MICA~\cite{bie2024mica} used multi-level cross-attention to align images with clinical concepts, generating concept localization maps via attention scores. CLAT~\cite{CLAT} encoded lesions with multiple learnable tokens in a transformer, and ExpLICD~\cite{Gao_Aligning_MICCAI2024} combined VLM-aligned textual criteria with visual concept tokens to produce heatmaps combining visual concept tokens with image feature maps.
Despite these advances, existing approaches still struggle to produce faithful and concept-discriminative visual localization maps. To address this, we propose a multi-concept token transformer that employs multiple learnable tokens, each targeted to represent a distinct concept. Additionally, our method provides visual explanations for each concept using solely image-level concept labels and without requiring additional supervision. To objectively evaluate the quality of these visual concept localization maps using quantitative metrics, ground-truth explanations or pixel-level annotations are often required, but they are rarely available in medical datasets. Moreover, the interpretation of clinical concepts can be challenging for non-experts. 
To support intuitive evaluation, we introduce \textit{SynSkin}, a synthetic dataset of over 10,000 skin lesion images annotated with six color labels and corresponding pixel-level masks. \textit{SynSkin} enables quantitative and qualitative assessment of explanations, supports multiple tasks such as multilabel classification and border detection, and allows both experts and non-experts to verify explanations without specialized training.

\section{Methodology}
\label{sec:methodology}

\subsection{Overview}
\begin{figure*}[!t]
    \centering
    \includegraphics[width=\linewidth]{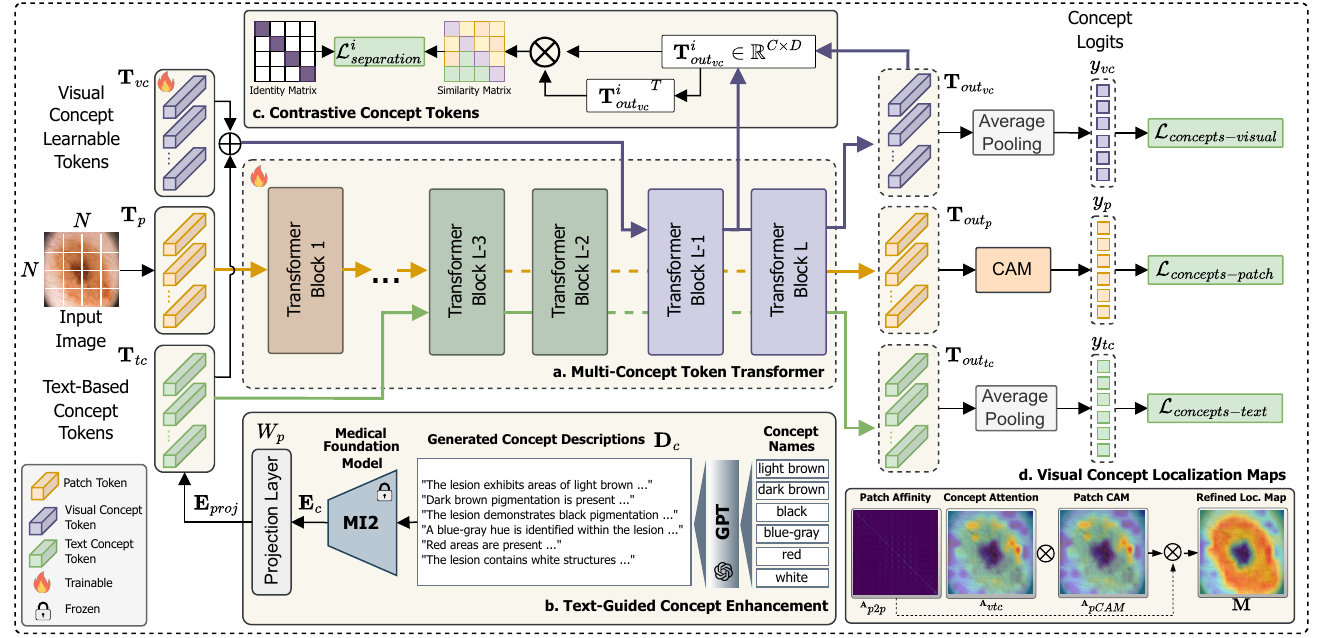}
    \caption{\textbf{Illustration of the proposed ViConEx-Med}. \textbf{(a.)} Multi-Concept Token Transformer encoder with specialized layers for processing multiple learnable visual concept tokens alongside complementary text-based concept tokens. \textbf{(b.)} Text-guided concept enhancement module, leveraging domain knowledge from a medical foundation model to provide complementary semantic information that guide the visual concept tokens. \textbf{(c.)} Contrastive concept token regularization applied to the output visual concept tokens to promote discriminativeness between visual tokens. \textbf{(d.)} At inference time, the visual concept localization maps are generated by fusing the multi-modal concept tokens with Patch CAM, followed by refinement using the pairwise affinity matrix derived from patch-to-patch attention.}
    \label{fig:viconex_architecture}
\end{figure*}

Medical imaging classification models typically operate on images paired with target labels to predict a diagnostic outcome directly. In this work, we reformulate the diagnostic task by first predicting a set of clinically relevant visual concepts that are commonly associated with the target diagnosis. For instance, the ABCDE criteria for melanoma detection~\cite{rigel2005abcde} emphasize features such as color variability within a lesion, where the presence of multiple distinct colors is clinically linked to malignancy. Such features are not only intuitive but also perceptible to non-experts. Accordingly, we formulate our task as a multilabel classification problem aimed at detecting the presence of clinical concepts in medical images. Given an image $x$, our model $f$ predicts $C$ concepts $\hat{y}_{cpt} = \sigma{(f(x))} \in [0,1]^C$, where $\sigma(.)$ is the sigmoid function. Beyond predicting the presence of each concept, our framework provides spatial localization of the corresponding regions by leveraging the attention mechanisms inherent to transformer-based architectures. Rather than offering a definitive diagnosis, our approach is positioned as a pre-diagnostic triage tool that categorizes lesions based on the identified visual concepts. 
Each module of ViConEx-Med is detailed in the following sections and visually depicted in Figure \ref{fig:viconex_architecture}.

\subsection{Multi-Concept Token Transformer}
Given an RGB image of $N \times N$ patches, these patches are fed into a patch embedding layer to produce a sequence of patch tokens, denoted as $\textbf{T}_p \in \mathbb{R}^{M\times D}$, where $D$ is the embedding dimension and $M = N^2$. 
To capture concept-specific information, we introduce $C$ learnable visual concept tokens, $\textbf{T}_{vc} \in \mathbb{R}^{C\times D}$, each targeted to represent a distinct clinical concept (e.g., color attributes). 
In parallel, we incorporate $C$ text-based concept tokens, $\textbf{T}_{tc} \in \mathbb{R}^{C\times D}$, initialized from textual embeddings of the corresponding concept semantics using a medical foundation model (\emph{c.f.} section \ref{subsec:text_guided_concept_module}). These text-based concept tokens are fed into the transformer encoder and also fused with visual concept tokens through element-wise summation, enriching each visual concept token with complementary semantic information from its textual counterpart.
A distinctive characteristic of our transformer encoder is the incorporation of specialized layers explicitly dedicated to visual and text concept attention. Given a transformer with $L$ layers, the first $l$ layers are designed for performing self-attention between image patches only. In the subsequent $m$ layers, the text-based concept tokens attend to the patch tokens to aggregate global information, while only the text-based tokens are updated during this stage. Finally, the last $n$ layers extract useful information from the set of processed patch embeddings to the visual concept-specific tokens, where $l+m+n=L$.
This decoupled design offers several benefits. By explicitly separating the transformer layers involving self-attention between patches from those dedicated to visual and text concept-attention, our model is able to progressively summarize relevant patch features into concept-specific representations. This enables the learning of enriched local contextual representations at multiple levels, resulting in more stable training and more discriminative concept features.

\subsubsection{Concept-aware training}
At the output of the transformer encoder, the output visual concept tokens $\textbf{T}_{out_{vc}} \in \mathbb{R}^{C\times D}$ are concatenated with the output patch tokens $\textbf{T}_{out_p} \in \mathbb{R}^{M\times D}$ and text concept tokens $\textbf{T}_{out_{tc}} \in \mathbb{R}^{C\times D}$. The combined sequence is then normalized to form the final output tokens of the transformer, $\textbf{T}_{out} \in \mathbb{R}^{(C+M+C)\times D}$. To ensure that each visual concept token captures unique and discriminative concept-related information, we apply channel-wise average pooling on the output visual concept tokens $\textbf{T}_{out}[1:C,:]$ to obtain the concept scores $y_{vc} \in \mathbb{R}^C$. These concept scores are supervised by the image-level ground-truth concept annotations $y \in \{0,1\}^C$ using the multilabel soft margin loss (MLSM):

\begin{equation}
\begin{split}
    \mathcal{L}_{concepts-visual} = \textrm{MLSM}(y_{vc}, y) = \\ 
    -\frac{1}{C} \sum_{i=1}^C y^i \log \sigma (y^i_{vc}) + (1-y^i) \log (1 - \sigma(y^i_{vc})).
\end{split}
\end{equation}

Similarly, we obtain the text-based concept scores $y_{tc} \in \mathbb{R}^C$ from the output text concept tokens by channel-wise averaging the output tokens $\textbf{T}_{out}[C:2C,:]$. Then, a MLSM loss is computed between the text concept scores and the ground-truth concept annotations:

\begin{equation}
\begin{split}
    \mathcal{L}_{concepts-text} = \textrm{MLSM}(y_{tc}, y).
\end{split}
\end{equation}

\begin{figure}[!t]
    \centering
    \includegraphics[width=\columnwidth]{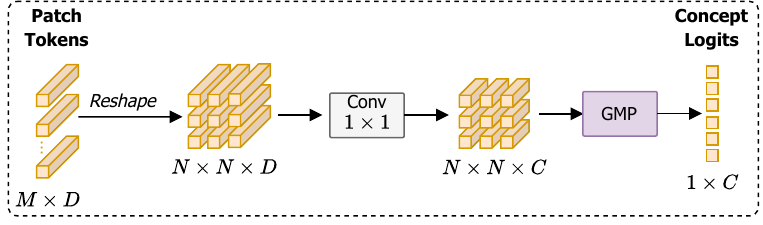}
    \caption{\textbf{The CAM module.} The output patch tokens from last layer of the transformer encoder are reshaped and then passed through a convolutional layer to reduce the depth to match the number of concepts. The resulting features are processed via Global Max Pooling to produce the concept scores.}
    \label{fig:cam_module}
\end{figure}

Although these multi-modal concept tokens capture rich information at the patch level, they are limited in extracting fine-grained local contextual information from the input image. To address this, we apply a Class Activation Mapping (CAM)~\cite{zhou2016learning} strategy on the output patch tokens, $\textbf{T}_{out}[2C:,:]$, to produce concept-specific scores. As illustrated in Figure \ref{fig:cam_module}, patch tokens are first reshaped and then projected into a 2D feature map $\textbf{F}_{out_{p}} \in \mathbb{R}^{N\times N \times C}$ using a convolutional layer with $C$ output channels. These feature maps are then processed through a Global Max Pooling (GMP) layer to produce the final concept scores $y_{p} \in \mathbb{R}^C$. The MLSM loss between the patch logits and the ground-truth concept annotations is given by:

\begin{equation}
\begin{split}
    \mathcal{L}_{concepts-patch} = \textrm{MLSM}(y_{p}, y).
\end{split}
\end{equation}

To obtain the final multi-modal concept predictions $\hat{y}_{cpt}$, we average all concept logits such that $\hat{y}_{cpt} = \sigma{(AVG(y_{vc},y_{p},y_{tc}))}$, where $AVG(.)$ denotes the arithmetic mean and $\sigma(.)$ corresponds to the sigmoid function.

\subsection{Text-Guided Concept Enhancement}
\label{subsec:text_guided_concept_module}

While visual concept tokens effectively capture discriminative and concept-specific visual information from the input image, they inherently lack the rich semantics of human language that clinicians use to describe and interpret these concepts. To bridge this gap, we complement the learnable visual concept tokens with text-based concept tokens derived from a medical imaging embedding model. More specifically, we leverage ChatGPT~\cite{OpenAI_ChatGPT} to generate medically informed $C$ textual descriptions for a predefined set of clinically relevant concepts (e.g., light-brown, dark-brown, black, blue-gray, red, and white). To maintain consistency with clinical terminology and ensure practical utility, we employ a carefully designed prompt tailored to elicit descriptions that mirror the tone and vocabulary used in clinical practice. 
The generated $C$ textual descriptions are subsequently encoded using the text encoder of MedImageInsight~\cite{codella2024medimageinsight}, an open-source medical imaging embedding model, denoted as $\Phi_{MI2}$. This yields $C$ textual embeddings:

\begin{equation}
    \textbf{E}_c = \Phi_{MI2} (\textbf{D}_c) \in \mathbb{R}^{C \times D_k},
\end{equation}

where $\textbf{D}_c$ denotes the textual descriptions, and $D_k$ is the dimensionality of the textual embedding space. To align these embeddings with the dimension of the learnable visual concept tokens, we introduce a projection layer $W_p \in \mathbb{R}^{D_k \times D}$:

\begin{equation}
    \textbf{E}_{proj} = W_pE_c \in \mathbb{R}^{C\times D},
\end{equation}

Finally, the text-based concept tokens are initialized as $\textbf{T}_{tc} = \textbf{E}_{proj}$, and kept constant during training. This ensures that these tokens preserve their original semantic grounding in clinical language, while providing complementary domain knowledge to guide the visual concept tokens.

\subsection{Contrastive Concept Token Regularization}
To enhance the discriminative capacity of the visual concept tokens, we incorporate a contrastive concept token regularization applied to the output visual concept tokens, inspired by~\cite{xu2024mctformerplus}. This regularization encourages the visual concept tokens to remain distinct from each other, promoting diversity in their attention over patch tokens. By enforcing separation in the learned representations, the model is driven to attend to different spatial regions for each concept, thereby improving both concept discrimination and localization performance. 
Specifically, as illustrated in Figure \ref{fig:viconex_architecture} (\textbf{c.}), given the output visual concept tokens $\mathbf{T}^i_{out_{vc}} \in \mathbb{R}^{C \times D}$ from the visual concept attention layer $i^{th}$, a pairwise similarity between two visual concept tokens is computed as $\textbf{S}^i = \mathbf{T}^i_{out_{vc}} \cdot ({\mathbf{T}^i_{out_{vc}}}^T)$, $\textbf{S}^i \in \mathbb{R}^{D \times D}$. To encourage each visual concept token to be similar to itself and dissimilar to all other visual concept tokens, a cross-entropy loss is applied between the similarity matrix $\textbf{S}^i$ and an identity matrix $\textbf{I} \in \mathbb{R}^{D \times D}$ as follows:

\begin{equation}
    \mathcal{L}^i_{separation} = \frac{1}{L} \sum_{i=1}^L CE(\textbf{S}^i, \textbf{I}),
\end{equation}

where $i$ is the index of the transformer encoding layer, $CE$ is the cross-entropy loss, and
$L$ is the number of the transformer encoding concept attention layers.

The total objective loss function for training the proposed ViConEx-Med is as follows:

\begin{equation}
\begin{split}
    \mathcal{L}_{total} = \alpha \mathcal{L}_{concepts-visual} + \beta \mathcal{L}_{concepts-patch} \\
    + \gamma \mathcal{L}_{concepts-text} + \delta \mathcal{L}_{separation},
\end{split}
\end{equation}

where $\alpha$, $\beta$, $\gamma$, and $\delta$ are hyperparameters controlling the weight of each loss term.

% Diagrama que mostra o processo de geracao do dataset
\begin{figure*}[!t]
    \centering
    \includegraphics[width=\linewidth]{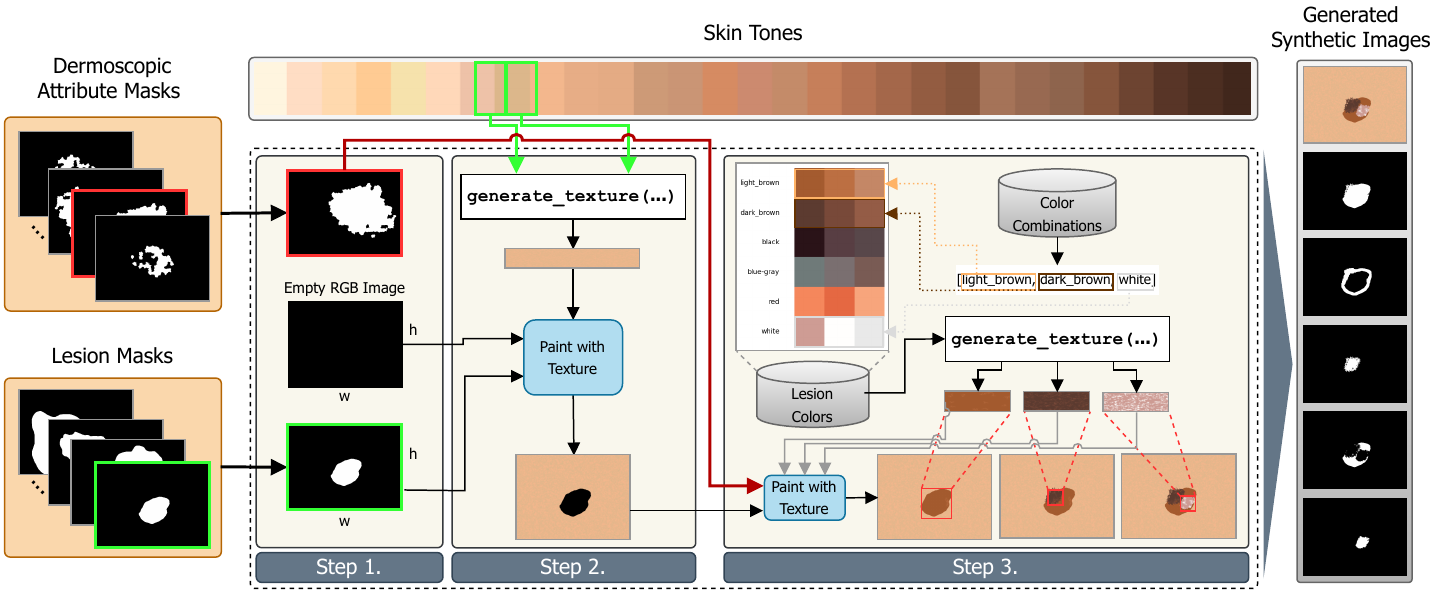}
    \caption{\textbf{Overview of the synthetic dataset generation pipeline}. \textit{Step 1}: A lesion mask and a randomly selected dermoscopic attribute mask are sampled from real skin datasets to serve as structural priors for the synthetic image. \textit{Step 2}: A background texture is generated using two consecutive sampled skin tones. \textit{Step 3}: A color combination is drawn from the color bank: the first color defines the lesion background color, while the remaining colors are applied to the attribute regions. The output includes the generated synthetic image along with the lesion mask, border mask and color masks.}
    \label{fig:synskin_generation_pipeline}
\end{figure*}

\subsection{Visual Concept Localization Maps}

At the inference stage, the attention maps for the visual concept tokens, $\textbf{A}_{vc} \in \mathbb{R}^{N\times N\times C}$, are obtained from the attention scores of the visual concept attention layers. Similarly, the text-based concept attention maps, $\textbf{A}_{tc} \in \mathbb{R}^{N\times N\times C}$, are extracted from the text concept attention layers. The multi-modal visual concept localization maps $\textbf{A}_{vtc}$ are computed via element-wise sum of these two maps:

\begin{equation}
    \textbf{A}_{vtc} = \textbf{A}_{vc} + \textbf{A}_{tc}. 
\end{equation}

Inspired by previous works~\cite{xu2021leveraging,xu2024mctformerplus}, we use patch CAM maps $\textbf{A}_{pCAM} \in \mathbb{R}^{N\times N\times C}$, derived from patch tokens (\emph{cf.} Figure \ref{fig:cam_module}), to complement the multi-modal visual concept localization maps $\textbf{A}_{vtc}$ via element-wise multiplication:

\begin{equation}
    \textbf{A} = \textbf{A}_{pCAM} \circ \textbf{A}_{vtc}. 
\end{equation}

The obtained concept localization maps $\textbf{A}$ can be further refined using a pairwise affinity matrix derived from patch-to-patch attention. More specifically, we extract the patch-to-patch attentions $\textbf{A}_{p2p} \in \mathbb{R}^{M \times M}$ from the self-attention layers of the transformer. The fused concept localization maps $\textbf{M} \in \mathbb{R}^{N\times N\times C}$ are given by:

\begin{equation}
    \textbf{M}(i,j,c) = \sum_{k=1}^{N} \sum_{l=1}^{N} \textbf{A}_{p2p}(i,j,k,l) \cdot \textbf{A}(k,l,c).
\end{equation}

The generated concept localization maps $\textbf{M}$ are then up-sampled to the original size of the input image before being normalized via min-max normalization.

\section{SynSkin Dataset}
\label{sec:synskin_dataset}

\subsection{Dataset Overview and Composition}

\begin{figure}[!t]
    \centering
    \includegraphics[width=\columnwidth]{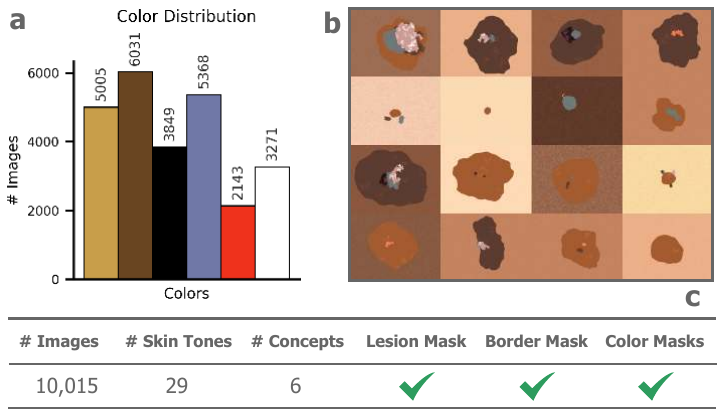}
    \caption{\textbf{Statistics for the SynSkin dataset.} \textbf{(a)} Color distribution. \textbf{(b)} Examples of synthetic images. \textbf{(c)} Statistics for SynSkin.}
    \label{fig:synskin_statistics}
\end{figure}

The \textbf{\textit{SynSkin}} dataset consists of 10,015 images, each annotated with six image-level color labels: \textit{red}, \textit{white}, \textit{blue-gray}, \textit{light-brown}, \textit{dark-brown}, and \textit{black}. These labels correspond to the clinically relevant colors described in the ABCDE criteria for melanoma detection~\cite{rigel2005abcde}. To simulate ethnic diversity, 29 distinct skin tones were incorporated during the image generation process. Each image is paired with binary masks indicating the lesion region, lesion border, and the presence of each annotated color. Figure \ref{fig:synskin_statistics} presents the statistics of the dataset and examples of synthetic images.

\subsection{Synthetic Data Generation Pipeline}

An overview of the pipeline for generating the synthetic dataset is illustrated in Figure \ref{fig:synskin_generation_pipeline}. 
The first step consists of randomly sampling a lesion segmentation mask and a dermoscopic attribute mask, which are drawn from real dermoscopic datasets to serve as structural priors to maintain realistic shape in the generated synthetic images. Specifically, the lesion segmentation masks are sampled from the HAM10000 dataset~\cite{tschandl2018ham10000}, while the dermoscopic attribute masks are sampled from the ISIC 2018 (Task 2) dataset~\cite{codella2019skin}.
To synthetically generate a visually plausible texture composed of distinct regions, we introduce a procedure based on randomized noise thresholding and morphological filtering. Given the selected lesion mask and an empty RGB image with the same dimensions as the lesion mask, we apply the texture to the background by choosing two consecutive skin tones from a set of predefined colors (\textit{cf.} Step 2. in Figure \ref{fig:synskin_generation_pipeline}). 
The final step involves randomly applying texture and color to the lesion regions. A color combination is first sampled from a predefined color set derived from the PH$^2$ dataset, where the color combinations were obtained by analyzing the frequency of clinically relevant colors to ensure consistency with real-world skin appearances. The first color in the selected combination corresponds to the base area of the lesion, defined by the initial segmentation mask, while the remaining colors are applied sequentially to create additional subregions within the lesion (\textit{cf.} Step 3. in Figure \ref{fig:synskin_generation_pipeline}). 
This procedure produces synthetic textures with well-defined, color-segmented regions of controlled size and complexity, making the generated images suitable for tasks such as semantic segmentation.
In addition to the generated synthetic image, we derive the corresponding segmentation masks to provide structural annotations. The final lesion mask is obtained by taking the union of all generated color region masks. 

% Figura a mostrar as estatísticas do dataset

\section{Experiments}
\label{sec:experiments}

%%%%%%%%%%%%%%%%%%%%%%%%%%%%%%%
% Tabela Estatisticas Datasets
%%%%%%%%%%%%%%%%%%%%%%%%%%%%%%%
\begin{table*}[!t]
    \caption{\textbf{Summary of datasets}. The distribution of color/dermoscopic/lesion labels in the six datasets. PN: Pigment Network; NN: Negative Network; STR: Streaks; MLC: Milia Like Cyst; GBL: Globules; EX: Exudates; HE: Hemorrhages; MA: Microaneuryms; SE: Soft Exudates.}
    \label{tab:dataset_statistics}
    \centering
    \setlength{\tabcolsep}{4pt}
    % Tabela PH2, Derm7pt, HAM10000 and SynSkin
    \begin{tabular}{lcccccccc}
        \toprule
        \multirow{2}{*}{\textbf{Dataset}} & \multirow{2}{*}{\textbf{Total}} & \multicolumn{6}{c}{\textbf{Color Labels}} & \multirow{2}{*}{\textbf{Train / Val / Test}} \\
        \cline{3-8}
        \noalign{\vspace{1mm}}
        & & Light-Brown & Dark-Brown & Black & Blue-Gray & Red & White & \\
        \midrule
        PH$^2$~\cite{PH2} & 200 & 139 & 156 & 42 & 38 & 10 & 19 & 158 / 21 / 21 \\
        Derm7pt~\cite{DERM7PT} & 827 & 372 & 485 & 108 & 101 & 51 & 18 & 346 / 161 / 320 \\
        HAM10000*~\cite{tschandl2018ham10000} & 3445 & 1557 & 2628 & 625 & 413 & 798 & 187 & 2646 / 599 / 200 \\
        \rowcolor{MyColorTab}
        \textbf{SynSkin} \footnotesize{(Ours)} & 10015 & 5005 & 6031 & 3849 & 5368 & 2143 & 3271 & 8014 / 998 / 1003 \\
        \bottomrule
    \end{tabular}
    % Tabela ISIC 2018
    \begin{tabular}{lccccccc}
        \toprule
        \multirow{2}{*}{\textbf{Dataset}} & \multirow{2}{*}{\textbf{Total}} & \multicolumn{5}{c}{\textbf{Dermoscopic Labels}} & \multirow{2}{*}{\textbf{Train / Val / Test}} \\
        \cline{3-7}
        \noalign{\vspace{1mm}}
        & & PN & NN & STR & MLC & GBL &  \\
        \midrule
        ISIC 2018~\cite{codella2019skin} & 3694 & 2244 & 275 & 183 & 773 & 828 & 2594 / 100 / 1000 \\
        \bottomrule
    \end{tabular}
    % Tabela DDR
    \begin{tabular}{lcccccc}
        \toprule
        \multirow{2}{*}{\textbf{Dataset}} & \multirow{2}{*}{\textbf{Total}} & \multicolumn{4}{c}{\textbf{Lesion Labels}} & \multirow{2}{*}{\textbf{Train / Val / Test}} \\
        \cline{3-6}
        \noalign{\vspace{1mm}}
        & & EX & HE & MA & SE &  \\
        \midrule
        DDR~\cite{li2019diagnostic} & 757 & 486 & 601 & 570 & 239 & 383 / 149 / 225 \\
        \bottomrule
    \end{tabular}
\end{table*}

\subsection{Experimental Settings}

\paragraph{Datasets.}
\label{subsec:datasets}
Experiments are performed on six datasets: PH$^2$~\cite{PH2}, Derm7pt~\cite{DERM7PT}, ISIC 2018~\cite{codella2019skin}, HAM10000~\cite{tschandl2018ham10000}, DDR~\cite{li2019diagnostic}, and the proposed \textbf{\textit{SynSkin}} dataset. The PH$^2$, Derm7pt, ISIC 2018, and HAM10000 datasets focus on the diagnosis of skin lesions and contain dermoscopic images of various melanocytic lesions, such as melanoma and nevus. In contrast, the DDR dataset targets the grading of diabetic retinopathy using retinal fundus images. The SynSkin dataset is synthetically generated and introduced in this work to enable controlled experiments and facilitate interpretability analysis.
We utilize PH$^2$, Derm7pt, HAM10000, and SynSkin for the task of color detection, targeting six clinically relevant colors: light-brown, dark-brown, black, blue-gray, red, and white. To evaluate the generalization of the proposed method beyond color detection task, we also employ ISIC 2018 and DDR datasets, which involve different clinical concepts, including dermoscopic attributes and retinal lesions, respectively.
Among these datasets, only PH$^2$ dataset provides image-level categorical color labels. To enable visual quality assessment, we generate the corresponding pixel-wise color masks for PH$^2$. For Derm7pt, we annotated both categorical and pixel-level color masks. For HAM10000, we restricted our experiments to the subset of images whose diagnosis was confirmed through histopathology, and manually annotated both categorical and pixel-wise color masks. For ISIC 2018~\cite{codella2019skin} dataset, we consider the lesion attribute detection subset, which provides pixel-level annotations of dermoscopic attributes. The DDR~\cite{li2019diagnostic} dataset comprises 13,673 retinal fundus images categorized into six diabetic retinopathy (DR) severity levels. In our experiments, we focus on the segmentation subset that includes 757 images with pixel-level annotations of retinal lesions.
A detailed overview of the datasets is provided in Table~\ref{tab:dataset_statistics}. All reported results (\emph{cf.} Section \ref{sec:results}) are obtained on the full test set of each dataset.

\paragraph{Implementation Details.}
\label{subsec:implementation_details}

We implemented the proposed ViConEx using CaiT-s24~\cite{touvron2021going} pre-trained on ImageNet~\cite{imagenet} as the backbone. We used AdamW optimizer with an initial learning rate of 4e-5 and a batch size of 32. For the text embeddings of concept descriptions, we employed the off-the-shelf MedImageInsight~\cite{codella2024medimageinsight} text encoder to extract 1024-dimensional embeddings.
For baseline comparisons, we employ ImageNet-pretrained ResNet50 and ViT Base models, fine-tuned on the concept prediction task. The remaining baseline methods are adapted to our task and retrained using publicly available implementations.
To evaluate the quality of the generated explanations, we employed the XAI metrics provided by the Quantus toolkit~\cite{hedstrom2023quantus}. Following the evaluation protocol in~\cite{rio2024suitability}, XAI metrics were computed only for correctly predicted concepts in each image of the test set.
All experiments were conducted on NVIDIA A6000 GPUs. To ensure robustness and reliability, results are reported as the average of three runs using different random seeds.

% Tabela color-based datasets
\begin{table*}[!t]
  \centering
  \small
  \caption{Comparison of our method against state-of-the-art models and CBMs on concept prediction and localization task across HAM10000, Derm7pt, PH$^2$, and \textbf{\textit{SynSkin}} datasets. Unit: \%. \includegraphics[height=1em]{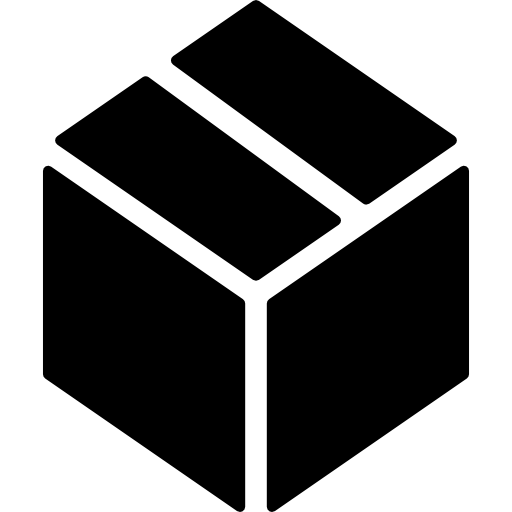} denotes a black-box model, while \includegraphics[height=1em]{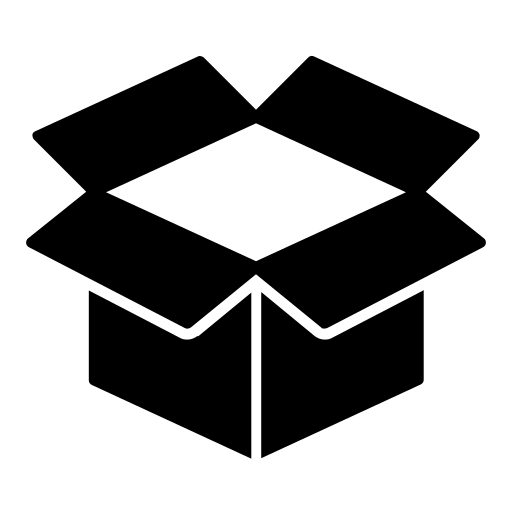} represents an interpretable model. The best results are highlighted in \textbf{bold}, and the
second-best results are \underline{underlined}. ACC: Accuracy. F1: F1-Score.}
  \label{tab:concept_prediction_color_datasets}
  \setlength{\tabcolsep}{4pt}
  \begin{adjustbox}{width=\linewidth}
  \begin{tabular}{llcccccccccccccccc}
    \toprule
     & \multirow{2}{*}{\textbf{Models}} & \multicolumn{4}{c}{\textbf{HAM10000}} & \multicolumn{4}{c}{\textbf{Derm7pt}} & \multicolumn{4}{c}{\textbf{PH$^2$}} & \multicolumn{4}{c}{\textbf{SynSkin}} \\
     \cmidrule(lr){3-6} \cmidrule(lr){7-10} \cmidrule(lr){11-14} \cmidrule(lr){15-18}
     & & ACC & AUC & F1 & Dice & ACC & AUC & F1 & Dice 
     & ACC & AUC & F1 & Dice & ACC & AUC & F1 & Dice \\
     \cmidrule(lr){1-2} \cmidrule(lr){3-6} \cmidrule(lr){7-10} \cmidrule(lr){11-14} \cmidrule(lr){15-18}
     \multirow{2}{*}{\includegraphics[height=1em]{images/closed-cardboard-box.png}} & ResNet-50 & 86.39 & 88.27 & 77.37 & - & 85.22 & 76.30 & 66.00 & - & 84.92 & 78.38 & 73.93 & - & 99.98 & 93.84 & 99.88 & - \\
     & ViT-base & 87.17 & 87.83 & 79.07 & - & 85.76	& 71.44 & 68.49 & - & 86.51 & 86.30 & 77.88 & - & 99.95 & 93.84 & 99.66 & -\\
     \cmidrule(lr){1-2} \cmidrule(lr){3-6} \cmidrule(lr){7-10} \cmidrule(lr){11-14} \cmidrule(lr){15-18}
     \multirow{5}{*}{\includegraphics[height=1em]{images/carton-box.png}} & CCBE \footnotesize{(CVPRW23)}~\cite{Patricio_2023_CVPRW} & 44.67 & 61.02 & 51.53 & 69.69 & 73.33 & 58.03 & 58.09 & 74.63 & 61.38 & 59.39 & 55.94 & 71.63 & 59.44 & 68.88 & 58.50 & 58.58 \\
     &MICA \footnotesize{(AAAI24)}~\cite{bie2024mica} & 47.17 & 50.10 & 52.01 & 60.94 & 49.08 & 44.32 & 47.16 & 66.18 & 57.41 & 50.13 & 56.48 & 62.95 & 45.65 & 50.43 & 60.65 & 49.75 \\
     &CLAT \footnotesize{(TMI24)}~\cite{CLAT} & 83.80 & 84.62 & 72.01 & \underline{73.12} & {81.70} & {67.28} & {57.99} & {72.32} & 85.18 & \textbf{72.43} & 74.99 & 66.07 & \textbf{99.85} & {99.99} & \textbf{99.82} & 60.59 \\
     &ExpLICD \footnotesize{(MICCAI24)}~\cite{Gao_Aligning_MICCAI2024} & 84.86 & 73.88 & 75.65 & 47.36 & 84.46 & 63.32 & 65.60 & 45.79 & 82.81 & 57.11 & 70.57 & 61.38 & {99.59} & {99.57} & {99.53} & 54.97 \\
     &MCTformer+ \footnotesize{(TPAMI24)}~\cite{xu2024mctformerplus} & {86.00} & \underline{87.94} & {76.42} & 69.81 & \underline{86.16} & \underline{75.61} & 69.04 & {76.01} & 84.65 & 58.58 & 76.26 & 70.63 & 99.75 & 99.64 & 99.70 & 57.78 \\
     \cmidrule(lr){1-2} \cmidrule(lr){3-6} \cmidrule(lr){7-10} \cmidrule(lr){11-14} \cmidrule(lr){15-18}
     \rowcolor{MyColorTab}
     & 
     \textbf{ViConEx-Med} \small{(Ours)} & \underline{86.33} & \textbf{88.05} & 77.60 & 72.20 & 86.10 & \textbf{76.62} & 68.96 & 75.94 & \textbf{88.36} & \underline{62.71} & \textbf{80.90} & 75.02 & \underline{99.79} & 99.99 & \underline{99.75} & \underline{63.59} \\
     \rowcolor{MyColorTab}
     & \quad \small{+ \textbf{\textit{SynSkin}} in the training set} & \textbf{86.47} & 87.13 & \underline{77.79} & \textbf{73.74} & \textbf{86.27} & 73.51 & \underline{70.18} & 77.05 & \underline{85.45} & 58.98 & \underline{77.13} & \underline{76.37} & - & - & - & - \\
     \rowcolor{MyColorTab}
     \includegraphics[height=1em]{images/carton-box.png} & \quad \small{+ \textbf{\textit{Visual-Text Token-Fusion}}} & \underline{86.33} & 87.88 & \textbf{78.00} & 72.43 & 86.08 & 75.73 & \textbf{70.31} & \underline{77.94} & 83.36 & 58.56 & 75.35 & 75.90 & 99.72 & \textbf{100.00} & 99.67 & 62.50 \\
     %\rowcolor{MyColorTab}
     %& \quad \small{+ \textbf{\textit{Text-Guided Concepts}}} & 85.17 & 83.16 & 75.86 & 72.41 & 85.57 & 70.12 & 69.02 & 77.11 & 84.12 & 58.85 & 74.68 & \textbf{78.95} & 99.77 & 99.99 & 99.73 & \underline{63.76} \\
     \rowcolor{MyColorTab}
     & \quad \small{+ \textbf{\textit{Hybrid}}} & 85.22 & 85.12 & 75.73 & 72.52 & 85.33 & 70.73 & 67.66 & \textbf{78.63} & 84.92 & 59.88 & 76.59 & \textbf{77.15} & 99.72 & 99.98 & 99.67 & \textbf{65.10} \\
     \bottomrule
  \end{tabular}
  \end{adjustbox}
\end{table*}

% Tabela com ISIC2018 e DDR
\begin{table}[!t]
  \centering
  \small
  \setlength{\tabcolsep}{3pt}
  \caption{Comparison of our method against state-of-the-art models and CBMs on concept prediction and localization task across ISIC 2018, and DDR datasets. Unit: \%. \includegraphics[height=1em]{images/closed-cardboard-box.png} denotes a black-box model, while \includegraphics[height=1em]{images/carton-box.png} represents an interpretable model. The best results are highlighted in \textbf{bold}, and the
second-best results are \underline{underlined}. ACC: Accuracy. F1: F1-Score.}
  \label{tab:isic2018_ddr_results}
  \begin{adjustbox}{width=\linewidth}
  \begin{tabular}{llcccccccc}
    \toprule
     & \multirow{2}{*}{\textbf{Models}} & \multicolumn{4}{c}{\textbf{ISIC 2018}} & \multicolumn{4}{c}{\textbf{DDR}} \\
     \cmidrule(lr){3-6} \cmidrule(lr){7-10}
     & & ACC & AUC & F1 & Dice & ACC & AUC & F1 & Dice \\
     \cmidrule(lr){1-2} \cmidrule(lr){3-6} \cmidrule(lr){7-10}
     \multirow{2}{*}{\includegraphics[height=1em]{images/closed-cardboard-box.png}} & ResNet-50 & 87.41 & 74.06 & 67.36 & - & 72.96 & 69.18 & 79.28 & - \\
     & ViT-base & 88.01 & 77.39 & 70.59 & - & 68.37 & 67.89 & 76.07 & - \\
     \cmidrule(lr){1-2} \cmidrule(lr){3-6} \cmidrule(lr){7-10}
     \multirow{5}{*}{\includegraphics[height=1em]{images/carton-box.png}} & CCBE \footnotesize{(CVPRW23)}~\cite{Patricio_2023_CVPRW} & 84.93 & 55.82 & 62.82 & 78.22 & 48.15 & 54.73 & 62.31 & 38.19 \\
     & MICA \footnotesize{(AAAI24)}~\cite{bie2024mica} & 45.33 & 45.71 & 41.74 & 68.99 & \textbf{74.59} & 45.66 & \textbf{80.45} & 33.96 \\
     & CLAT \footnotesize{(TMI24)}~\cite{CLAT} & 86.07 & 67.59 & 64.77 & 73.53 & \underline{73.52} & \underline{68.93} & \underline{79.39} & 25.00 \\
     & ExpLICD \footnotesize{(MICCAI24)}~\cite{Gao_Aligning_MICCAI2024} & 73.41 & 52.44 & 50.91 & 77.23 & 68.22 & 52.76 & 75.21 & 28.67 \\
     & MCTformer+ \footnotesize{(TPAMI24)}~\cite{xu2024mctformerplus} & {87.85} & {77.35} & {69.74} & {78.04} & 67.59 & 68.39 & 76.27 & 33.91 \\
     \cmidrule(lr){1-2} \cmidrule(lr){3-6} \cmidrule(lr){7-10}
      \rowcolor{MyColorTab}
    &  \textbf{ViConEx-Med} \small{(Ours)} & 88.05 & 74.79 & \underline{70.52} & \textbf{79.91} & 72.63 & 68.68 & 78.82 & 37.62 \\
     \rowcolor{MyColorTab}
     \includegraphics[height=1em]{images/carton-box.png} & \quad \small{+ \textbf{\textit{Visual-Text Token-Fusion}}} & 88.05 & \underline{77.64} & 70.42 & 78.07 & 71.63 & \textbf{69.50} & 78.60 & \underline{38.43} \\
     %\rowcolor{MyColorTab}
     %\includegraphics[height=1em]{images/carton-box.png} & \quad \small{+ \textbf{\textit{Text-Guided Concepts}}} & \textbf{88.28} & 77.59 & \textbf{71.83} & 79.14 & 70.37 & \textbf{69.70} & 77.74 & \underline{38.62} \\
     \rowcolor{MyColorTab}
     & \quad \small{+ \textbf{\textit{Hybrid}}} & \textbf{88.18} & \textbf{77.93} & \textbf{71.26} & \underline{79.47} & 70.30 & {67.35} & 77.63 & \textbf{40.20} \\
     \bottomrule
  \end{tabular}
  \end{adjustbox}
\end{table}

\paragraph{Evaluation Metrics.}
\label{subsec:evaluation_metrics}

To evaluate the performance of the proposed method, we employ Accuracy, F1-score and AUC. For the task of concept localization, we use Dice score as the evaluation metric. The Dice score is calculated as $\frac{2|\textbf{A} \cap \textbf{B}|}{|\textbf{A}| + |\textbf{B}|}$, where $\textbf{A}$ represents the model prediction and $\textbf{B}$ denotes the ground truth. To simultaneously assess classification performance and localization quality, we define the Classification-Localization Score (CL-Score) as $\sqrt{F1 \cdot Dice}$. To further evaluate the quality of the generated visual concept explanations, we adopt XAI metrics that assess common properties of explainability, including faithfulness, robustness, complexity, and localization~\cite{rio2024suitability}. Specifically, we measure Selectivity~\cite{montavon2018methods}: how the model’s predictions change when features with the highest attribution values are removed; Sparseness~\cite{chalasani2020concise}: whether the explanations are concentrated on a small set of features or dispersed across many features; Pointing Game~\cite{zhang2018top}: whether the location of the highest attribution falls within the ground-truth object region; and Continuity~\cite{montavon2018methods}: the sensitivity of the explanations to small perturbations in the input image.

\section{Results}
\label{sec:results}

\subsection{Comparison with State-of-the-Art Methods}

We compare our method and its variants against recent state-of-the-art approaches on the tasks of concept prediction and localization across multiple datasets. Specifically, we consider the following configurations: \textbf{ViConEx-Med} is the baseline architecture including only the visual concept tokens. \textbf{ViConEx-Med + Visual-Text Token-Fusion} extends the baseline by fusing the text-based concept tokens with the visual concept tokens. In this variant, the text-based concept tokens are not fed into the transformer encoder. \textbf{ViConEx-Med + Hybrid} represents the full proposed method, including all modules and components, as illustrated in Figure \ref{fig:viconex_architecture}.

\paragraph{Concept Prediction Performance.}

\begin{figure}[!t]
    \centering
    \includegraphics[width=\columnwidth]{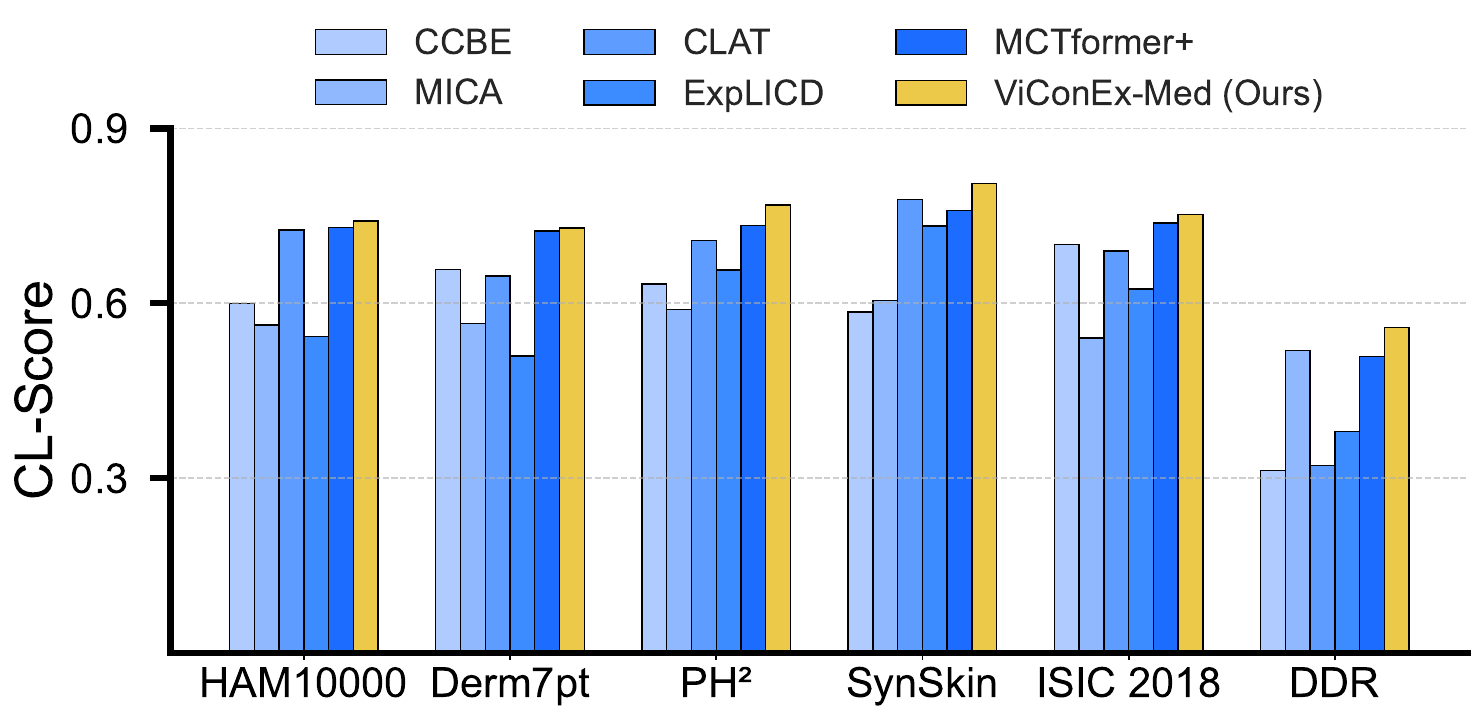}
    \caption{\textbf{Comparison of CL-Score (range $[0, 1]$) across six datasets for the evaluated methods.} Higher CL-Score indicates better alignment between concept predictions and localized regions. Our method (yellow bar) consistently outperforms baseline approaches in all datasets.}
    \label{fig:clscore}
\end{figure}

% Color attributes datasets
For datasets where the target concepts correspond to color attributes, we report the results in Table \ref{tab:concept_prediction_color_datasets}. Our baseline method achieves comparable, and in several cases superior, performance relative to existing approaches. Among the competing approaches, CLAT, ExpLICD, and MCTformer+ achieve comparable results, consistent with their use of transformer architectures and cross-attention mechanisms for concept-level alignment. In contrast, CCBE and MICA yield the lowest performance. The weaker results of CCBE can be attributed to the poor discriminability of the encoder and the oversimplification of pooling in the bottleneck layer for concept prediction, while MICA also relies on CNN-based backbone and depends on a rich concept set to learn concept activation vectors (CAVs)~\cite{kim2018interpretability}, constraining its capacity for end-to-end optimization. When augmenting the training set with the SynSkin dataset, our method improves over the baseline variant and achieves the best performance on HAM10000 and Derm7pt dataset in terms of accuracy. Notably, ViConEx-Med also matches or outperforms black-box models, while providing visual concept explanations.
% Clinical concepts datasets
Complementarily, for datasets where the target concepts involve clinical concepts beyond color attributes, ViConEx-Med also achieves competitive or superior performance compared to existing methods (Table \ref{tab:isic2018_ddr_results}). Among these, MICA and CLAT yield the best detection performance on the DDR dataset, with MICA attaining the highest accuracy and F1-score but a substantially lower AUC (45.66\%), while CLAT performs well given its design specifically optimized for retinal disease diagnosis. Notably, when incorporating text-based concept guidance into our method, ViConEx-Med Hybrid achieves the highest CL-Score compared to current approaches, as illustrated in Figure \ref{fig:clscore}. These results highlight the ability of our method to generalize effectively across both color-based and clinical concepts on concept prediction task.

\begin{figure}[!t]
    \centering
    \includegraphics[angle=90,width=\columnwidth]{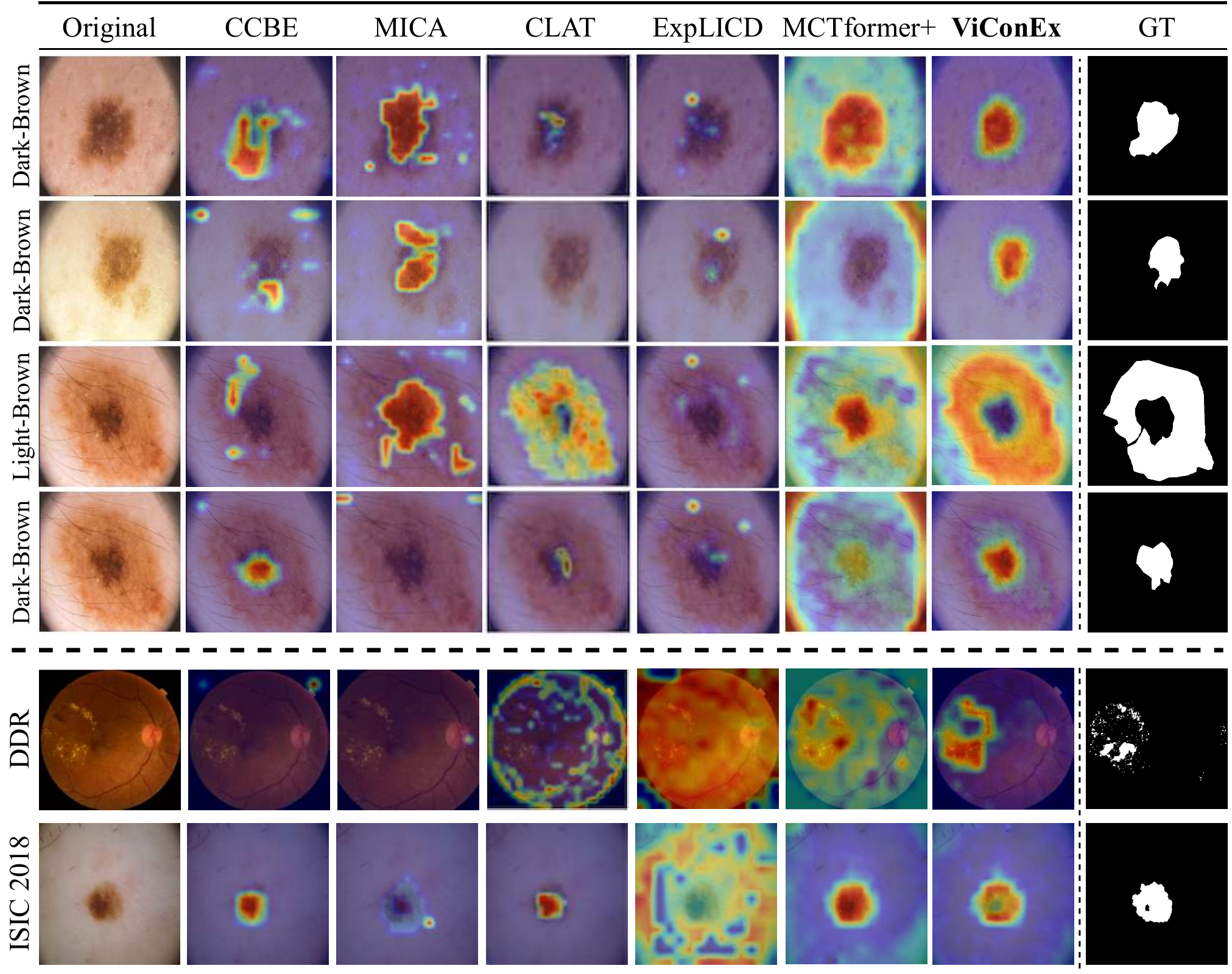}
    \caption{\textbf{Qualitative comparison of the concept localization maps by different methods}. \textit{Top:} color localization maps. \textit{Bottom:} clinical concept localiation maps. ViConEx yields a more focused and discriminative region.}
    \label{fig:qualitative_results_color_datasets}
\end{figure}

\paragraph{Concept Localization Performance.}
We evaluate concept localization performance using the Dice score. As shown in Tables \ref{tab:concept_prediction_color_datasets} and \ref{tab:isic2018_ddr_results}, ViConEx-Med consistently outperforms existing methods across datasets. In particular, the hybrid variant of ViConEx-Med achieves the best or second-best Dice score in 5 out of 6 datasets. Augmenting the training set with SynSkin further improves Dice scores in HAM10000, Derm7pt and PH$^2$, highlighting the utility of synthetic data for enhancing the localization performance. Consistent with these findings, Figure \ref{fig:clscore} demonstrates that our method excels in terms of joint classification-localization trade-off (CL-Score). Moreover, qualitative comparisons (Figure \ref{fig:qualitative_results_color_datasets}) reveal that ViConEx-Med consistently produces more focused and discriminative localized regions that better align with the ground-truth masks compared to competing approaches. These results confirm the effectiveness of ViConEx-Med in providing accurate concept predictions while achieving precise localization across diverse datasets. 

% Figure CL-Scores

% Figure color localization maps

\paragraph{Quality of Visual Concept Explanations.}
We assess the quality of visual concept explanations using XAI metrics capturing different dimensions of interpretability, including faithfulness, robustness, complexity, and localization. The results of ViConEx-Med are compared with the second-best competing method (MCTformer+) in Table \ref{tab:xai_metrics}. Notably, ViConEx-Med (Hybrid) achieves the best results across all metrics. 
Specifically, a lower Selectivity indicates higher faithfulness, as sequentially removing features with the highest attribution values leads to a stronger performance drop in classification accuracy. A higher Sparseness score reflects more robust and focused attributions, measured via the Gini index. A higher Pointing Game score indicates better localization, as the attributions with highest values falls within the object regions of the target mask. Finally, lower Continuity reflects more stable explanations, as the attribution maps vary less under input perturbations.
These results show that ViConEx-Med not only improves predictive performance but also provides more faithful, robust, and interpretable explanations.

% Table XAI metrics
\begin{table}[!t]
  \centering
  \caption{Comparison of XAI metrics averaged across datasets. The best results are highlighted in \textbf{bold}, and the
second-best results are \underline{underlined}. Select.: Selectivity. Sparse.: Sparseness. P. Game: Pointing Game. Contin.: Continuity.}
  \label{tab:xai_metrics}
  \begin{adjustbox}{width=\columnwidth}
  \begin{tabular}{lcccc}
    \toprule
     \textbf{Models} & \textbf{Select.} $\downarrow$ & \textbf{Sparse.} $\uparrow$ & \textbf{P. Game} $\uparrow$ & \textbf{Contin.} $\downarrow$ \\
     \cmidrule{1-5}
     \multicolumn{5}{c}{\textit{All Datasets}} \\
     \cmidrule{1-5}
     MCTformer+ \footnotesize{(TPAMI24)}~\cite{xu2024mctformerplus} & 0.5294 & 0.0467 & 0.3213 & 42.10 \\
     \cmidrule(lr){1-1} \cmidrule(lr){2-5}
     \rowcolor{MyColorTab}
     \textbf{ViConEx-Med} \footnotesize{(Baseline)} & {0.4448} & {0.2034} & {0.4150} & {33.08} \\
     \rowcolor{MyColorTab}
     \textbf{ViConEx-Med} \footnotesize{(Visual-Text Token-Fusion)} & 0.4220 & 0.2070 & 0.4893 & 34.29 \\
     \rowcolor{MyColorTab}
     \textbf{ViConEx-Med} \footnotesize{(Text-Guided Concepts)} & \underline{0.4142} & \underline{0.2268} & \underline{0.5253} & \underline{31.75} \\
     \rowcolor{MyColorTab}
     \textbf{ViConEx-Med} \footnotesize{(Hybrid)} & \textbf{0.3929} & \textbf{0.2295} & \textbf{0.5341} & \textbf{29.38} \\
     \cmidrule{1-5}
     \multicolumn{5}{c}{\textit{Color Detection Datasets}} \\
     \cmidrule{1-5}
     MCTformer+ \footnotesize{(TPAMI24)}~\cite{xu2024mctformerplus} & 0.4572 &	0.1993&	0.5835	&33.70 \\
     \cmidrule(lr){1-1} \cmidrule(lr){2-5}
     \rowcolor{MyColorTab}
     \textbf{ViConEx-Med} \footnotesize{(Baseline)} & \underline{0.3757}  &	0.2093 &	0.6467 &	30.13 \\
     \rowcolor{MyColorTab}
     \textbf{ViConEx-Med} \footnotesize{(Baseline+SynSkin)} & 0.3803 &	\underline{0.2146} &	\underline{0.6502} &	\underline{27.79} \\
     \rowcolor{MyColorTab}
     \textbf{ViConEx-Med} \footnotesize{(Hybrid)} & \textbf{0.3729}	& \textbf{0.2364} &	\textbf{0.6907}	 &\textbf{26.55} \\
     \bottomrule
  \end{tabular}
  \end{adjustbox}
\end{table}

\begin{table}[!t]
  \centering
  \caption{Evaluation of the effectiveness of each regularization term and pooling strategy on the \textbf{\textit{SynSkin}} dataset. Unit: \%. The best results are highlighted in \textbf{bold}, and the
second-best results are \underline{underlined}.}
  \label{tab:ablation_losses}
  \begin{adjustbox}{width=0.85\columnwidth}
  \begin{tabular}{lcccc}
    \toprule
     \textbf{Pooling} & $\mathcal{L}_{con-patch}$ & $\mathcal{L}_{con-visual}$ & $\mathcal{L}_{con-mean}$ & {\textbf{Dice}} \\
     \cmidrule(lr){1-1} \cmidrule(lr){2-4} \cmidrule(lr){5-5}
     GMP & \xmark & \xmark & \cmark & 62.91 \\
     GMP & \cmark & \cmark & \xmark & \textbf{63.59} \\
     GMP & \cmark & \cmark & \cmark & 61.37 \\
     \cmidrule(lr){1-1} \cmidrule(lr){2-4} \cmidrule(lr){5-5}
     GAP & \xmark & \xmark & \cmark &  61.57 \\
     GAP & \cmark & \cmark & \xmark & 62.89 \\
     GAP & \cmark & \cmark & \cmark & 62.57 \\
     \cmidrule(lr){1-1} \cmidrule(lr){2-4} \cmidrule(lr){5-5}
     GWRP & \xmark & \xmark & \cmark & 61.41 \\
     GWRP & \cmark & \cmark & \xmark & \underline{63.53} \\
     GWRP & \cmark & \cmark & \cmark & {63.35} \\
     \bottomrule
  \end{tabular}
  \end{adjustbox}
\end{table}

\subsection{Ablation Studies on ViConEx-Med}
\label{subsec:ablation_studies}

\begin{figure}[!t]
    \centering
    \includegraphics[width=0.85\columnwidth]{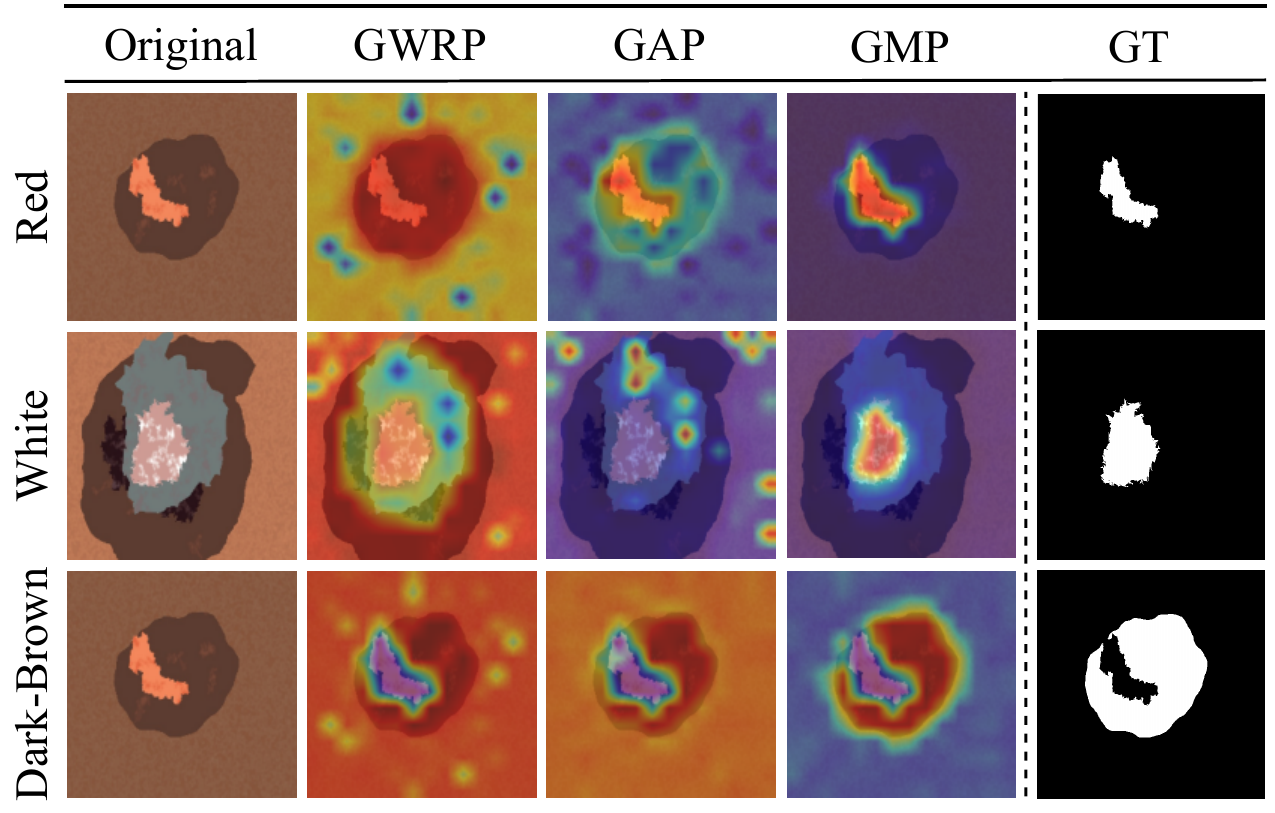}
    \caption{\textbf{Examples of color localization maps under different pooling strategies on SynSkin dataset.} With GMP the localization maps are more focused and discriminative.}
    \label{fig:ablation_pooling_synskin}
\end{figure}

\paragraph{Effect of loss function terms and pooling strategies.} The CAM module in the ViConEx-Med architecture (\emph{cf.} Figure \ref{fig:cam_module}) supports multiple pooling strategies for aggregating patch tokens, including Global Average Pooling  (GAP), Global Max Pooling (GMP), and Global Weighted Ranking Pooling (GWRP)~\cite{xu2024mctformerplus}. The method of combining concept logits, either by a unified loss (mean concept logits), or by separate losses for each set of predicted concepts affects classification performance. The results of Table \ref{tab:ablation_losses} on the SynSkin dataset show that GMP with separate losses for $\mathcal{L}_{con-patch}$ and $\mathcal{L}_{con-visual}$ achieves the best Dice score (63.59), followed by GWRP employing the same configuration. Overall, we observe that employing separate losses consistently outperforms both the mean and global loss strategies. Visualizations in Figure \ref{fig:ablation_pooling_synskin} confirm that GMP produces more localized and discriminative regions aligned with ground-truth, motivating its adoption as the default pooling strategy.

\paragraph{Effect of multimodal tokens in ViConEx-Med.} As shown in Tables \ref{tab:concept_prediction_color_datasets}, \ref{tab:isic2018_ddr_results}, and \ref{tab:xai_metrics}, incorporating text-based concept guidance improves concept localization and prediction. While simple fusion of semantic information with visual concept tokens (ViConEx-Med  Visual-Text Token-Fusion) slightly decreases CL-Score (-0.24\%), introducing specialized attention layers for text-based concept tokens in the transformer encoder (ViConEx-Med Hybrid) yields an average improvement of +0.22\%. In contrast, incorporating text-based concept tokens without fusion (ViConEx-Med Text-Guided Concepts) results in a minor decrease (-0.09\%). These findings underscore the importance of specialized attention mechanisms for effectively complementing visual concept representations with semantic context.

\section{Conclusion}
\label{sec:conclusions}

In this paper, we introduce ViConEx-Med, a multi-concept token transformer for interpretable concept prediction, designed to produce visual concept localization maps associated with each predicted concept. Through extensive experiments on six medical datasets, we demonstrated that our method outperforms state-of-the-art methods in both concept prediction and localization. Moreover, our method offers competitive performance with black-box approaches, while offering faithful explanations to the predicted concepts. This effectiveness stems from its decoupled transformer-based architecture, which separates the processing of visual and text-based concept tokens from patch-level self-attention, enabling richer contextual representations. Validation with XAI metrics confirms that the produced visual concept localization maps are both faithful and robust, offering a transparent and human-understandable decision-making process, which is critical in clinical settings. Beyond performance, ViConEx-Med offers a plug-and-play framework that supports human-in-the-loop and seamless integration with emerging LVLMs, paving the way for safer and trustworthy deployment of AI models in healthcare.

\section*{Acknowledgements} This work was funded by the Portuguese Foundation for Science and Technology (FCT) under the PhD grant ``2022.11566.BD'' (\url{https://doi.org/10.54499/2022.11566.BD}), and supported by UID/04516/NOVA Laboratory for Computer Science and Informatics (NOVA LINCS) with the financial support of FCT.IP.

% References
\bibliographystyle{IEEEtran}
\bibliography{main}

\end{document}